\documentclass[]{llncs}
\usepackage[utf8]{inputenc}
\usepackage[english]{babel}
\usepackage{amsmath}
\usepackage{amsfonts}
\usepackage{amssymb}
\usepackage{graphicx}
\usepackage{hyperref}
\usepackage{color}
\usepackage{caption}
\newcommand{\ssc}{\emph{SSCSP}}
\newcommand{\ssca}{\emph{SSCSP-Solution}}
\newcommand{\sscb}{\emph{structured SSCSP-Solution}}

\newcommand{\bfs}{\emph{BFS}}

\newcommand{\ff}[1]{{#1}}

\newcommand{\lift}{\mathit{lift }}
\begin{document}

\author{Ulrich Furbach \inst{1} \and  Florian Furbach \inst{2}  \and Christian Freksa \inst{3} }
\institute{University Koblenz \email{uli@uni-koblenz.de}
\and TU Kaiserslautern \email{furbach@cs.uni-kl.de}
\and University of Bremen \email{freksa@uni-bremen.de}}


\date{}

\title{Relating Strong Spatial Cognition to Symbolic Problem Solving --- An Example}

\maketitle

\begin{abstract}
In this note, we discuss and analyse a shortest path finding approach using strong spatial cognition. It is  compared  with a symbolic graph-based algorithm and it is shown that both approaches are similar with respect to structure and complexity. Nevertheless, the strong spatial cognition solution is easy to understand and even pops up immediately when one has to solve the problem.
\end{abstract}

\section{Introduction}
In discussions among cognitive scientists over the past 25 years or so about the undisputed merit of graphic depictions for problem solving, the use of classical formal analysis of graphic information processing occasionally caused some discomfort. Firstly, because  intuitively graphical procedures often appear simpler than their formal counterparts,  and secondly, because  even highly accomplished theoreticians emphasize the usefulness of graphics for getting insight into suitable problem-solving approaches~\cite{polya2014solve}. Therefore it appears inappropriate to some researchers to analyse graphic information processing by means of an informationally but not structurally equivalent formal representation.

While debates about the complexity of graphic information processing focused on the use of static spatial information in graphics, the strong spatial cognition (SSC) paradigm goes one step further and addresses spatial transformation of spatial configurations for problem solving, i.e. replacing computational operations by physical operations.

\begin{figure}[h]

\centering
\includegraphics[scale=0.3]{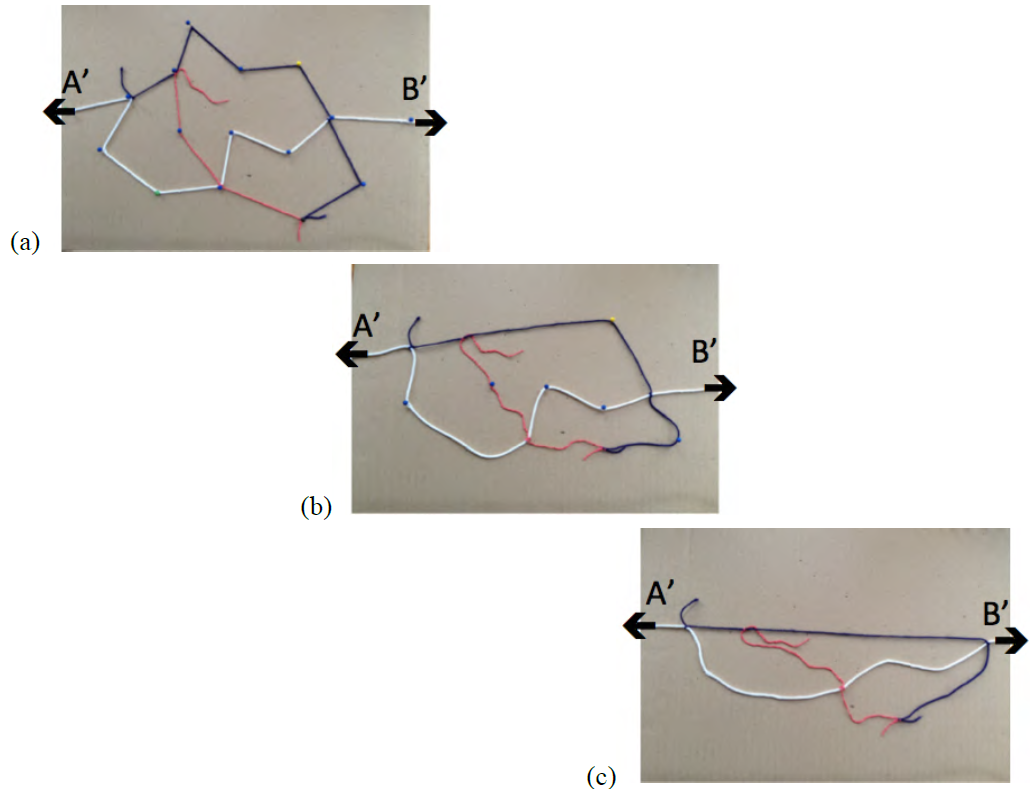} 
\caption{Mild abstraction of a route network (from \cite{DBLP:conf/cosit/Freksa15}).
Pulling apart strings at A’ and B’ distorts angles and shapes of the route network but preserves the relative distances in (b) and (c). 
The shortest route is the straight connection between A’ and B’ in (c).}
\label{freksa}

\end{figure}

 Such problems have in common that parts of the problem are real world objects, instead of representations within a computer. 
In \cite{DBLP:conf/cosit/Freksa15}
 such an example is discussed:  a  map, where parts of the reality, e.g. streets or places are represented by real world objects, namely drawings consisting of lines and polygons on paper.  Such a representation is called a weak or mild abstraction  i.e. an abstraction where properties of objects and relations, which are relevant for solving certain problems are preserved. In \cite{DBLP:conf/ki/FreksaFD84} the notion of intrinsic properties of parts of a representation was used to describe a similar effect. In the map example these are the relative length and orientation of streets and the size of a place. This can be used to solve problems like localization or path finding. An alternative to a map on paper would be an abstraction of the real world, where routes are represented by strings and locations by connections between the strings.  An example is depicted in Figure~\ref{freksa}a, which we will use later on in order to analyse a path finding problem. 

\begin{figure}%
\centering
\includegraphics[width=.6\linewidth]{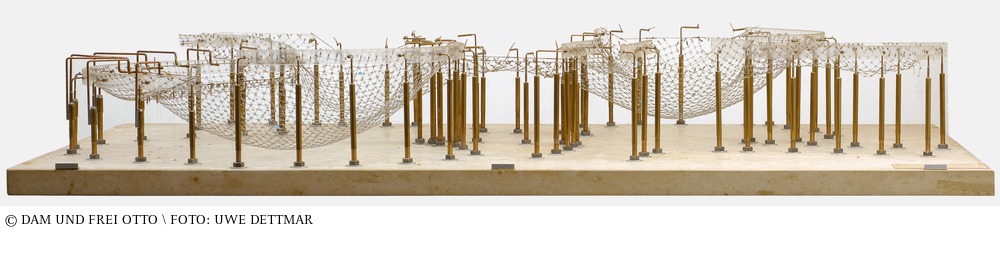}%
\hspace{0.5cm}
  \includegraphics[width=.3\linewidth]{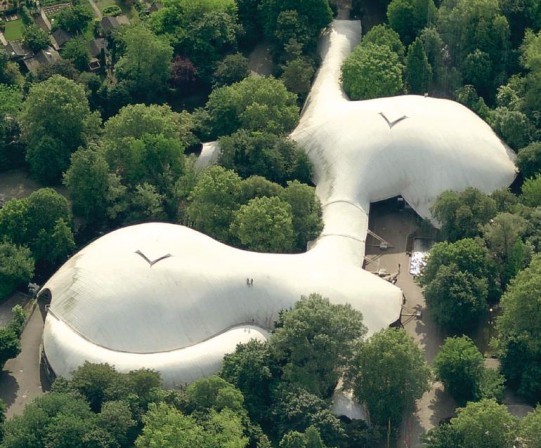}
\caption{On the left is the hanging chain model of the Multihalle in Mannheim on  display in the Deutsches Architektur Museum. 
On the right is the upside-down reality version. \\Source: Deutsches Architektur Museum and Stadt Mannheim.
}%
\label{fig:mannheim}%
\end{figure}

Another famous example for using mild abstractions for solving real world problems is the hanging chain method used to model catenary curves. This is used e.g. by the Catalan architect Gaudi for the construction of organic designs of churches, like the Sagrada Familia. Gaudi used weighted strings to determine the optimal static design of arcs in his architecture, instead of computing the curves, whose math was well known at this time (due to work of Leibniz, Huygens and Bernoulli on these catenary curves).
%
This techniques is used in modern architecture as well. As an  example  the  'Multihalle Mannheim' together with its chain model is shown in Figure~\ref{fig:mannheim}

In spatial cognition research there is increasing work, which aims 
at developing models that take into account the role of the body and the spatial environment. The interaction between mind, body, and environment helps to find solutions to spatial problems. As an example, we will analyse in  this paper a  shortest path solution which is based on the mild abstraction as given in Figure~\ref{freksa}, this is called the \ssc-Problem  (\underline{S}trong \underline{S}patial \underline{C}ognition \underline{S}hortest \underline{P}ath) in the following.

The next section introduces the {\ssca}; in order to analyse this problem solving method we will \ff{simplify it and }transform it in a subsequent section to a symbolic algorithm\ff{, the \sscb , which  will be be analysed and compared with graph algorithms}. Since symbolic algorithms usually are analysed by complexity considerations, we will try to apply a complexity measure to the \sscb\ as well.
\medskip

\section{\ssc\ in a Physical World}
We will now use strings as a mild abstraction as illustrated in Figure~\ref{freksa}. Here, we use strings to represent routes between two locations. The intersections of those strings represent the intersections in a route network. The length of the strings are given by a fixed scaling factor  of the corresponding routes. So the relative lengths of the routes are preserved. {The task is is to find the shortest path between two locations $A$ and $B$ within a route network. This is the \ssc-Problem.
We now use this string representation in order to solve it.
 The locations $A$ and $B$ have representations $A'$ and $B'$  in the string model.} In order to find the shortest path between $A$ and $B$ one simply has to pull apart strings at $A'$ and $B'$ ---  voil\`a.
It is immediately clear that the shortest route is given by those strings that are pulled straight and the length of the shortest path is represented by the distance between $A'$ and $B'$. \ff{This is the \ssca .}
Interestingly, we managed to solve the problem without analysing the network in any way, all we had to do was to pull. 

The complexity seems to be constant --- you need one operation 'pulling apart'. 
In contrast, if we represent the network by a graph using a data structure in a computer, we know that the complexity of the problem is exponential in the number of nodes  on the path.


It is astonishing how humans manage to solve the shortest path problem immediately by just pulling the strings. And even more: they are convinced that this yields a shortest path --- this property is just 'popping up'.
This motivates us to analyze this solution within the mild abstraction paradigm in more detail, putting special  emphasis on the complexity aspect.

 In computer science we focus on information and we are interested in the complexity of algorithms wrt.\ information processing effort required in terms of processing steps and storage capacity. We do not take into account mass and energy when we discuss complexity of information processing. But how can we assess the complexity of problem-solving processes appropriately when we manipulate information by other means than in classical telecommunication or computer architectures, where bits of information are propagated through static or ad hoc networks? To our knowledge, there is no specific theory for information processing complexity of graphic information processing such as in depictive geometry, scientific graphs, or geographic maps where some of the operations may consist of manipulating physical and spatial entities. 

We are interested in  the question of whether the complexity of solving a problem is inherent only in the problem and its information-theoretical representation and processing structures or whether it also can be a function of the physical and spatial substrate in which the problem is manifested.

In the following section we will transform the \sscb\ into a graph-based algorithm, namely breath-first search ---  a common  approach in symbolic problem solving. Furthermore, we propose an approach for comparing complexities of both methods. We will see that both methods, {the \sscb\ and the graph-search} have a very similar structure --- the astonishing effect is that one needs not to be aware of this structure, when the problem is solved using the mild abstraction. Once the problem is transformed into a symbolic graph-based representation, one needs to know about algorithms for graph-search to solve the problem.

\section{From \ssc\ to a symbolic algorithm}
We start by analysing a simplified version of the problem:  We assume that all strings connecting nodes have the same length $d$.
Furthermore, we will only pull on node $A'$ instead of both, $A'$ and $B'$.
Now, we observe the pull operation and notice that the closer a node is to $A'$, the earlier it starts to move.

\begin{figure}%
\centering
\includegraphics[width=.3\linewidth]{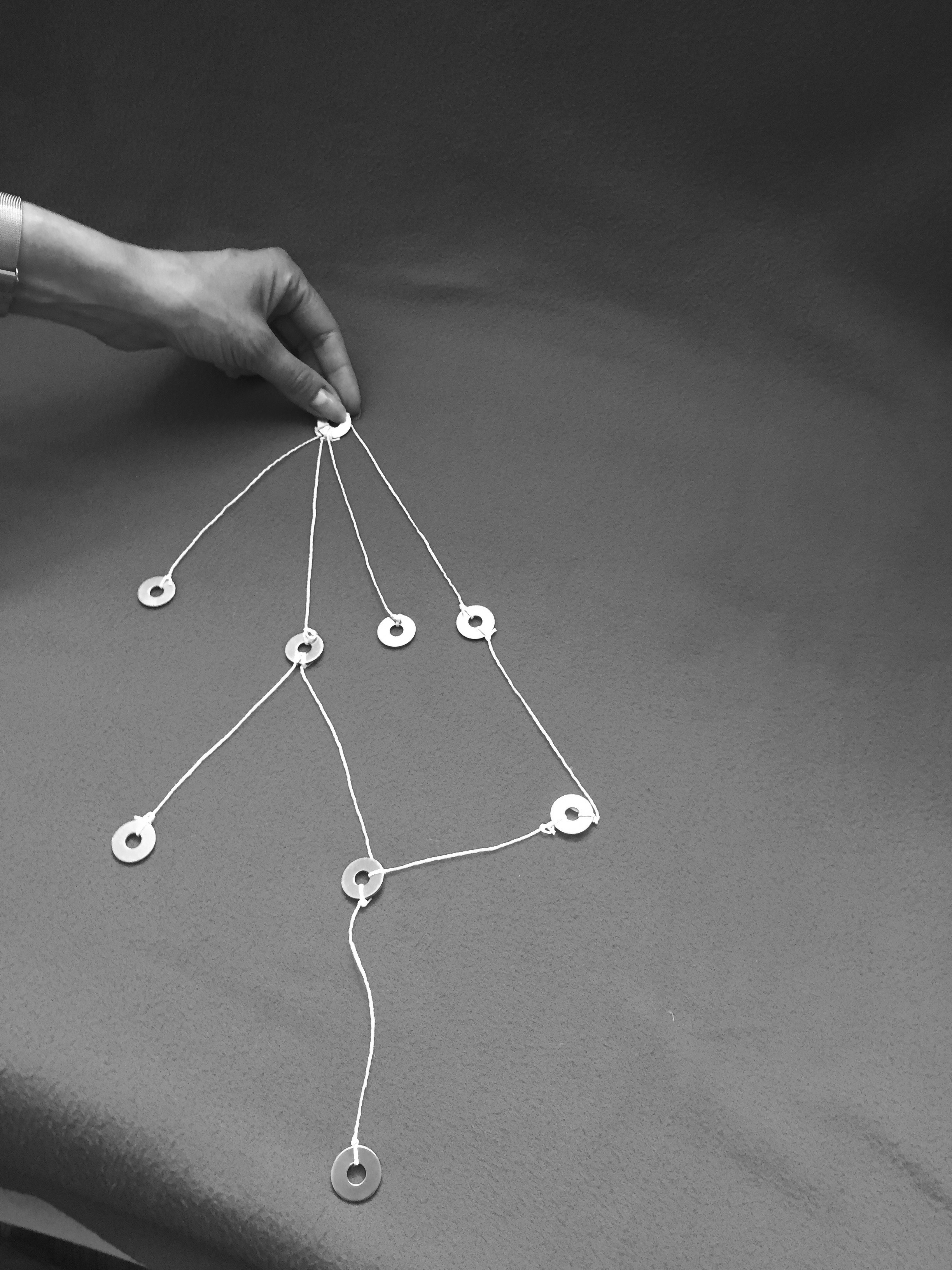}%
\hspace{0.25cm}
  \includegraphics[width=.3\linewidth]{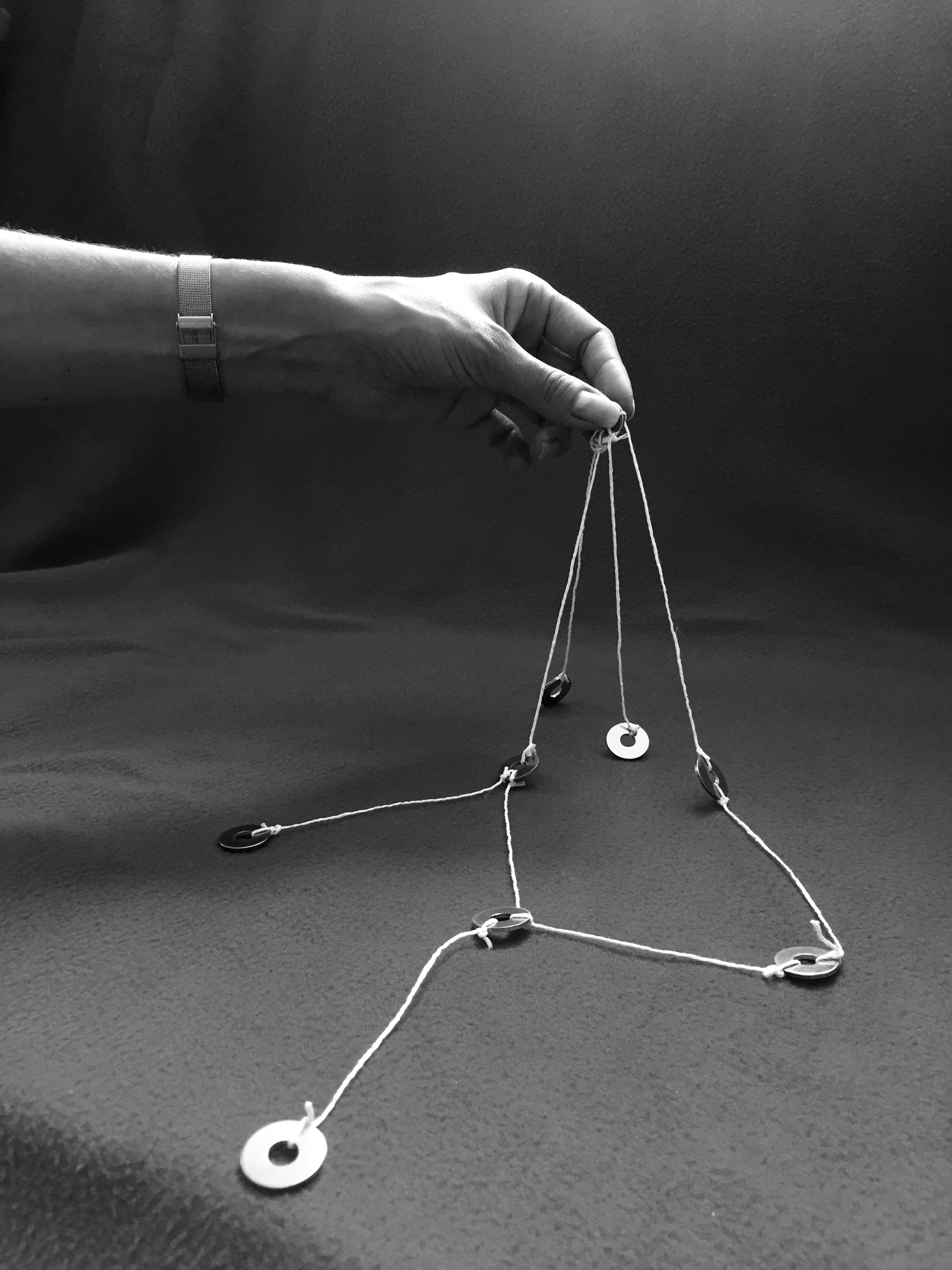}
 \hspace{0.25cm}
  \includegraphics[width=.3\linewidth]{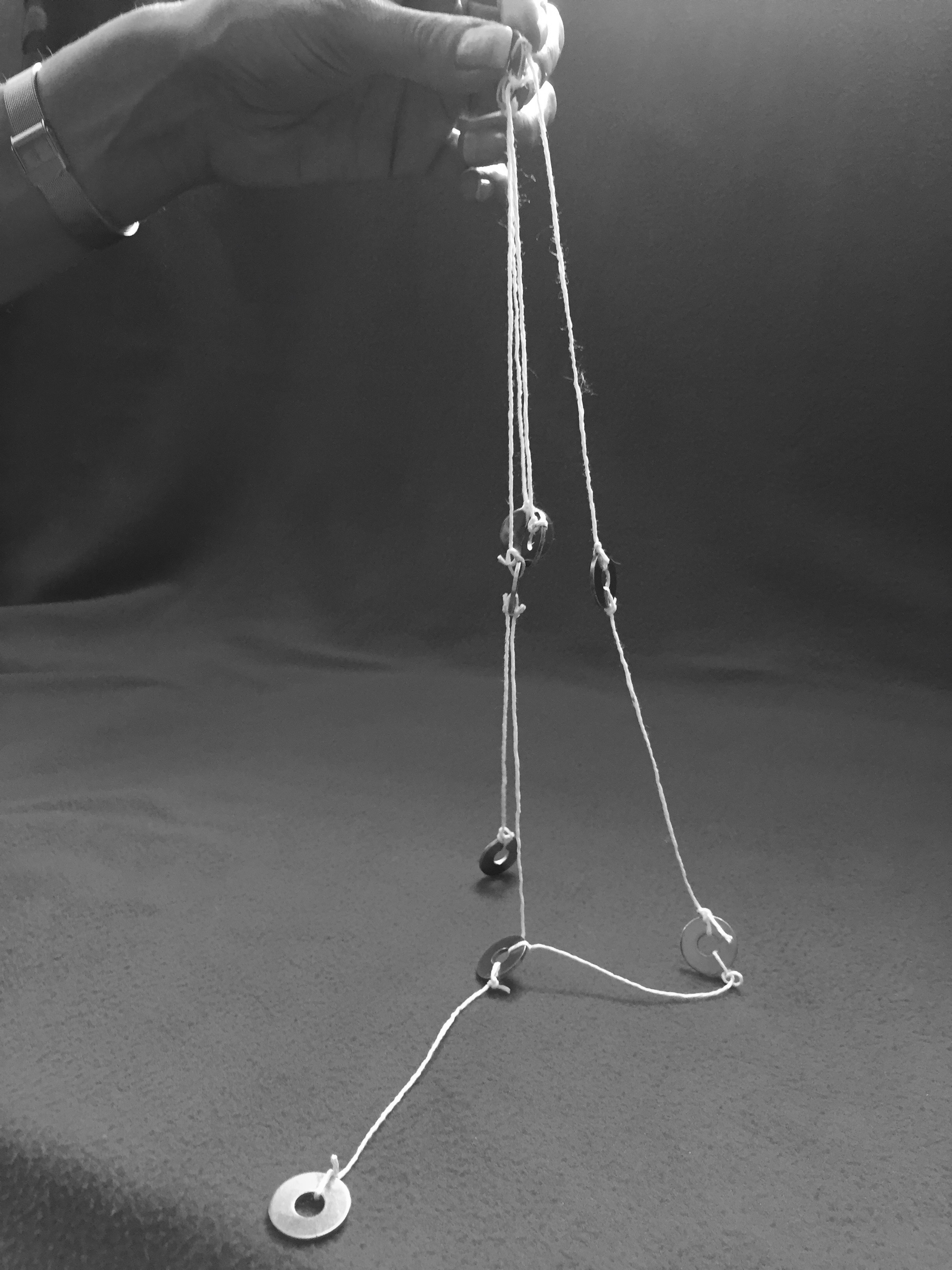} 
\caption{Mild abstraction -- structured version. Left: Network is spread on a flat surface. One node is selected by grab-operation. Middle: Grabbed node is lifted until nodes from the first layer are ready to left from the surface. Right: Lifting proceeds until nodes from layer 3 are ready to be lifted.}%
\label{fig:graph}%
\end{figure}

In order to get this more precisely, we change our model. We assume the network is spread on a flat surface. Instead of pulling nodes apart, we assume an operation grab, which selects {only one} node and the pulling apart is changed into a lift of the grabbed node vertically into the third dimension until $B'$ is lifted from the surface {as well}. {Once $B'$ is lifted, the shortest path is given by the sequence of strings by which $B'$ hangs from $A'$.}
Now, we can decompose the lift into single steps, where we only pull up until a new node is lifted from the surface (this is depicted in Figure~\ref{fig:graph}). With respect to the work needed to lift nodes we see easily that lifting the nodes into the third dimension certainly can be seen as an upper bound compared to the work necessary to pull the nodes apart in two dimensions.
This enables us to describe our solution to shortest path in a more detailed way than before.

{\bf The \sscb} is  given by:
\begin{itemize}
\item grab node $A'$
\item repeat until node $B'$ is lifted: 
\begin{itemize}
\item[] increase  height of $A'$ over surface by distance $d$ 
\end{itemize} 
\end{itemize}

 Note that, if a node $x$ is lifted from the surface, the next lifting operation causes all adjacent nodes of $x$ still on the surface to be lifted together. This is another simplification of our model; depending on the structure of the network it may happen that during one lift-operation an adjacent node is lifted earlier than its neighbors. This may occur e.g. if one node has no adjacent node, while its neighbor is connected to larger part of the network. The connecting string to the latter will  form a kind of hanging chain curve, because of the friction of the part of network, which is still on the surface. This node is lifted after a distance smaller than $d$, whereas  the other node, the one  without a further connection, will be lifted only at the end of the lift operation by distance $d$. This effect is ignored in the following -- we simply assume that there is no friction between nodes and the surface. Thus work is only necessary for lifting nodes.

With these assumptions, we can  construct a formal method of solving the shortest path problem for uniform edge length in a theoretical representation as an undirected graph. We represent the lifting of nodes simply by adding nodes to a set $\lift$--- in one lift operation all adjacent nodes of a node are lifted at the same time. This corresponds to the insertion of all adjacent nodes in one step into the set $\lift$. The nodes in the set $\lift$ correspond to those nodes which are lifted above the surface.
\bigskip

{\bf As graph search} we  have the following algorithm:

\begin{itemize}
\item $\lift = \{ A'\}$
\item repeat until node $B' \in \lift$ 
\begin{itemize}
\item[] for each element in $\lift$ add all its adjacent nodes to $\lift$
\end{itemize} 
\end{itemize}

This is actually a breadth first search algorithm (\bfs)\cite{5219222}.
{The distance between $A$ and $B$ is given by the number of loop iterations required and the shortest path can simply be derived with backtracking. In order to keep it simple, we have not included this in our description.}
This version of \bfs\ appears to be rather naive: it does not include a marking of nodes which already have been expanded --- there is an 'attempt' for every node to add its adjacent nodes into $\lift$ even if this has been done already in previous steps.

\section{Comparing both Approaches} 
Despite the apparent similarity of both approaches, the time complexities are rather different. The time of the \sscb\  is linear in the distance between $A'$ and $B'$ and the time of \bfs\ depends on the structure of the graph. 
Comparing the two approaches shows immediately that in iteration $i$ of the \sscb\  all nodes {that are $i$ nodes or less away} from $A'$ are lifted {by $d$}. They are all lifted simultaneously.
Whereas in \bfs\ in iteration $i$ all nodes distance $i$ or less are added to $\lift$. The run-time of this depends on the implementation and the architecture available. The \sscb\ can be seen as a kind of parallel version of \bfs .
However, if there are sufficiently many parallel processes available and 
with the appropriate implementation, \bfs\ can have a similar run-time.

Time complexity of graph algorithms usually is measured by counting the number of nodes which have to be visited during search. 
Instead of counting nodes, we base our complexity analysis on the {physical }work necessary to perform the lift operation of nodes.
For the complexity of the \sscb\ we now assume all nodes have a weight $w$ and the strings are weightless.  It requires a work of $w\cdot d$ to lift a node.  For simplification, let us assume that the grab operation requires also work $w\cdot d$. Let $l(i)$ be the number of nodes with distance {(i.e. number of nodes on shortest path)} $i$ from $A'$. Let the $n$ be the distance from $A$ to $B$. {A node with distance $i$ from $A'$ is lifted $(n-i+1)$ times.} The work required by the \sscb\ is 
\begin{equation}
w \cdot d \cdot \sum_{i=0}^n (n-i+1)\cdot l(i). \label{ssp} 
\end{equation}

For the complexity of \bfs\ we now assume that adding nodes requires a time $t$ for each node in the resulting set $\lift$. For simplification, let us assume that the initialization of lift requires time $t$.
The time required by \bfs\ is 
\begin{equation}
t \cdot \sum_{i=0}^n (n-i+1)\cdot l(i).  \label{bfs}
\end{equation}

Comparing Equations \ref{ssp} and \ref{bfs} shows immediately  that the work of \ssc\ directly corresponds to the running time of \bfs.

However, in common implementations of \bfs, the run-time is not only dependent on how many nodes are added. 
Rather, we pick each node $x$ in lift and then add all its adjacent nodes consecutively. 
This often results in nodes being added multiple times in a single iteration. 
Assume now that each such node addition requires time $t$.  
Let $N(i)$ be the set of nodes that have distance $i$ to $A'$. Let $g(x)$ be the out--degree of a node $x$. 
Now, the time required by \bfs\ is 
\begin{equation}
			 t \cdot \sum_{i=0}^{n-1}\left( (n-i)\cdot\sum_{x\in N(i)} g(x)\right) .
\end{equation}

This complexity also transfers to the \sscb\ if we assume that each string has a weight $w$ and the nodes are weightless. 
Since the nodes are weightless the grab operation requires no work and lifting a node $x$ by $d$ requires work $g(x)\cdot w$.\footnote{The work to lift a string from the surface is obviously smaller than $g(x)\cdot w$, because the string which corresponds to an edge is not lifted at once in entirety. Moreover, this is done in a continuous way resulting in less work. Hence, Equation \ref{over} is an over estimation.}
The work required by the \sscb\ is 
\begin{equation}
\label{over}
				 w\cdot d \cdot \sum_{i=0}^{n-1}\left( (n-i)\cdot\sum_{x\in N(i)} g(x)\right) .
\end{equation}


The \bfs\ is usually accelerated. In each iteration, it is enough to add only nodes adjacent to nodes newly added in the previous iteration.
Here, we could show that the complexity is similar to the \ssc-Solution if we assume that only lifting a string off the surface requires work. This, however, seems to be an unrealistic assumption; the previously lifted nodes have to be also lifted, which of course, needs more work. This is an example for an optimization of \bfs\ which does not carry over to \ssc.

If we do not require uniform string length between nodes, the parallels become less clear. The most prominent approach here is Dijkstra's shortest path algorithm\cite{ref1} which processes the nodes and edges in the order of their distance to the starting node. This is similar to the \ff{\ssc-Solution}. However, there does not seem to be a direct correspondence between the run-time of an execution of Dijkstra and the \ssc-Solution.

We would like to emphasize again, that the complexity consideration in Equation~\ref{over} was done mainly to analyse the structure of the \ssc-Solution and to compare it with the symbolic \bfs. We are  aware, that the representation of a problem by using mild abstractions allows very well an adaption of the physical objects  to the power available for the solution. E.g. in the \ssc-Problem from Figure~\ref{freksa} the entire layout of the graph and the lenghth of the strings is such, that we can do the pulling apart very easily - without beeing aware of the 'work-complexity' defined above.

\section{Conclusion}

In this note we presented and analyzed a shortest path procedure using strong spatial cognition. We compared the approach with a symbolic graph search algorithm by estimating the amount of physical work required to construct a shortest path. We arrived at the conclusion that structure and complexity of both approaches can easily be mapped to one another. This is surprising as the solution of the problem when presented with mild abstraction, i.e. by employing physical objects, pops up immediately; we do not become aware of the internal structure and the physical complexity of the approach. In comparison, the symbolic graph-based algorithm can be designed and applied only with knowledge about data and graphs available.

What is the reason for our perception of different complexity of the symbolic vs. the physical problem solving approach to the shortest path problem? In the symbolic approach, we compose algorithms from structurally simple elements, using knowledge about the length of path segments and explicitly combining these elements to essentially building up the network of paths, keeping track of those parts of the network that promise to yield the shortest overall path. In comparison, in the SSC approach we do not deal with components; we do not even have to know any length to identify a shortest path. The approach exploits intrinsic spatial relations that are implicit in the network of strings and cannot be violated. For this reason it will be difficult to program the approach incorrectly. Physical affordances and constraints do the work for us and we do not have to understand how the strings and nodes move for the approach to yield a correct result.

\bibliographystyle{plain}
\bibliography{Freksa15}

\end{document}